\documentclass[10pt,twocolumn,letterpaper]{article}

 \usepackage[pagenumbers]{cvpr} 

\usepackage[dvipsnames]{xcolor}

\definecolor{cvprblue}{rgb}{0.21,0.49,0.74}
\usepackage[pagebackref,breaklinks,colorlinks,citecolor=cvprblue]{hyperref}

\def\paperID{} 
\def\confName{CVPR}
\def\confYear{2024}

\title{DragTex: Generative Point-Based Texture Editing on 3D Mesh}
\author{Yudi Zhang\quad\quad\quad Qi Xu\quad\quad\quad Lei Zhang \\
Beijing Institute of Technology\\
{\tt\small therealyudiz@gmail.com} \quad {\tt\small qpril9757@gmail.com} \quad
{\tt\small leizhang@bit.edu.cn}
}

\usepackage{makecell}
\begin{document}

\twocolumn[{%
\renewcommand\twocolumn[1][]{#1}%
\maketitle

\begin{center}
    \centering
    \captionsetup{type=figure}
    \includegraphics[width=\textwidth]{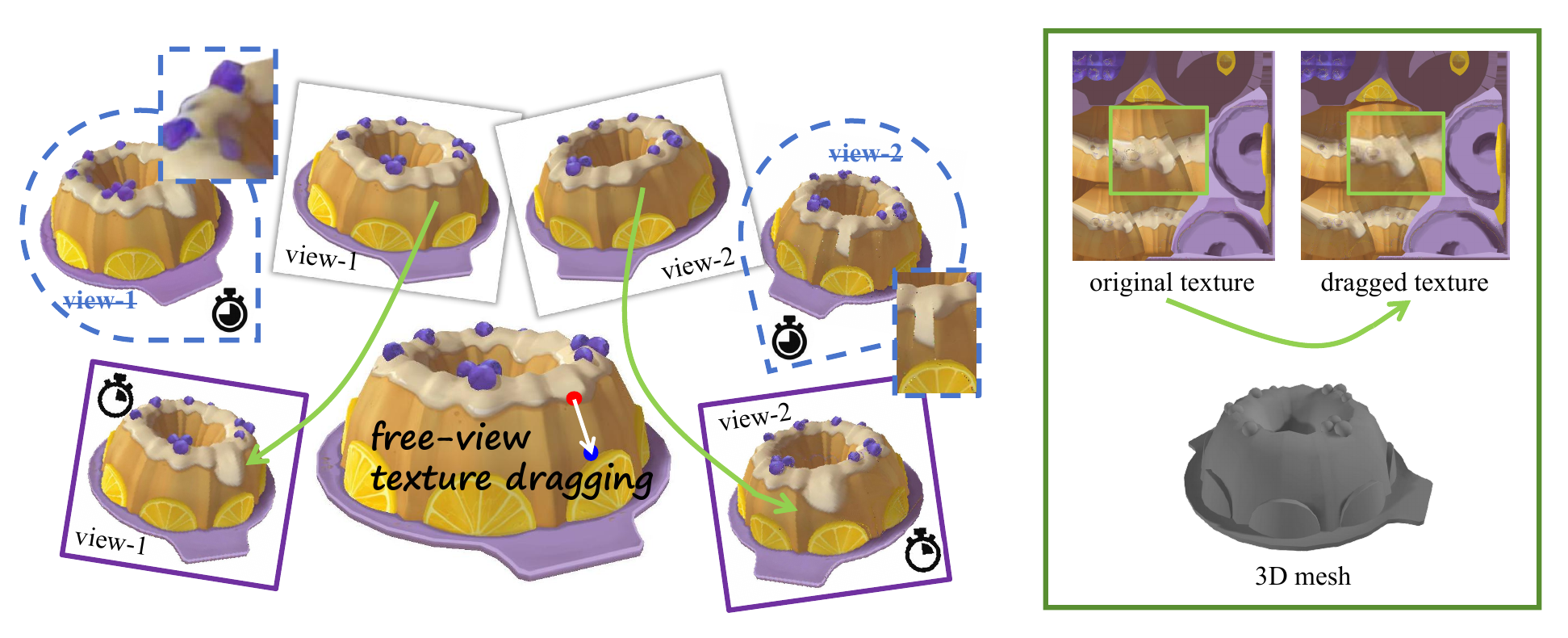}
    \vspace{-15pt}
    \captionof{figure}{Within a freely selected view, DragTex facilitates an intuitive drag interaction directly on a textured 3D mesh (indicated by the white arrow), producing a new texture that aligns with the intended drag movement and maintains consistency across multiple views (see images enclosed by purple boxes) instead of the texture generated with the naive method (see images enclosed by blue dashed boxes).}
    \label{Fig:teaser}
\end{center}%
}]

\begin{abstract}
Creating 3D textured meshes using generative artificial intelligence has garnered significant attention recently. While existing methods support text-based generative texture generation or editing on 3D meshes, they often struggle to precisely control pixels of texture images through more intuitive interaction. While 2D images can be edited generatively using drag interaction, applying this type of methods directly to 3D mesh textures still leads to issues such as the lack of local consistency among multiple views, error accumulation and long training times.
To address these challenges, we propose a generative point-based 3D mesh texture editing method called DragTex. This method utilizes a diffusion model to blend locally inconsistent textures in the region near the deformed silhouette between different views, enabling locally consistent texture editing. Besides, we fine-tune a decoder to reduce reconstruction errors in the non-drag region, thereby mitigating overall error accumulation. Moreover, we train LoRA using multi-view images instead of training each view individually, which significantly shortens the training time. The experimental results show that our method effectively achieves dragging textures on 3D meshes and generates plausible textures that align with the desired intent of drag interaction.
\end{abstract}

\section{Introduction}
\label{sec:intro}

The mesh with texture is a representative form of 3D content (see Figure~\ref{Fig:teaser}), which is widely used in computer animation, games, virtual reality, \emph{etc}. The recent advancements in Artificial Intelligence Generated Content (AIGC) have opened up new avenues for generative 3D content creation \cite{gao2022get3d,metzer2023latent,raj2023dreambooth3d,poole2022dreamfusion}. For example, interactive editing of textures on 3D meshes allows users to design and refine generated assets through text-based or local scribble-based manipulations \cite{richardson2023texture}, offering great flexibility. Nevertheless, it is inadequate to use these types of editing to precisely guide the spatial change of textures to align with the user's intent during interaction.

Some existing methods \cite{pan2023drag,shi2023dragdiffusion,ling2023freedrag,mou2023dragondiffusion} have successfully enabled point-based dragging interaction for generative editing in the context of 2D images. However, the extension of these techniques to the manipulation of textures on the surface of a 3D mesh poses several challenges. Firstly, directly mapping a dragged image from a selected view back to a texture atlas can lead to noticeable seams in neighboring views. These seams result from the deformation of textures near the objects' silhouettes between views, leading to inconsistencies in the local texture region (\eg, the image of view-2 in the blue dashed box in Figure~\ref{Fig:teaser}). Secondly, the generative drag editing process, including denoising and decoding, tends to introduce artifacts in regions where edits do not overlap (\eg, the image of view-1 in the blue dashed box in Figure~\ref{Fig:teaser}). These artifacts arise due to the decoder's limited ability to accurately reconstruct unedited regions, which subsequently impacts drag editing results of views that encompass these distorted regions, thus diminishing the overall quality of the texture. This problem is what we call error accumulation. Thirdly, the direct application of 2D image drag editing techniques like DragDiffusion \cite{shi2023dragdiffusion} requires separate training for each view, resulting in an increased demand for both training time and computational resources for the editing process.

To address the aforementioned issues, we adapt the interactive generation of 2D images to 3D mesh textures, towards generating textures that are not only plausible but also consistently aligning with the intended edits. This is accomplished by enabling direct point-based dragging on the surface of the 3D mesh through the so-called \emph{DragTex}. We draw inspiration from methods of single-view image drag editing, while blending the noisy latent images originating from both the regions that have been dragged and those that have not, particularly in the vicinity near the silhouette of the object. Such blending ensures a uniform and coherent representation of textures across views, thus eliminating the issue of noticeable seams and inconsistent local texture deformation (see Figure \ref{Fig:teaser}). Furthermore, a fine-tuning strategy can be applied to guide the decoder to simultaneously preserve the effects of the drag interaction in the edited region and the intricate details of the original texture in the untouched regions. An additional enhancement involves the utilization of multi-view images of the textured mesh. Instead of training a single-view low-rank adaption model (LoRA \cite{hu2021lora}) before each drag interaction, we opt to pre-train a LoRA model with multi-view images before commencing any point-based drag. This strategic alteration is to reduce the training time.

The main contribution of this work is a generative editing method tailored for texture images on 3D meshes, also the first point-based texture editing method. It empowers users to perform point-based dragging interactions directly on the mesh surface. To achieve this, we propose noisy latent image blending and the fine-tuning of the decoder to enhance the local consistency and overall quality of textures. 
For the training process, we employ a multi-view LoRA training approach, strategically chosen to enhance efficiency. The experimental results demonstrate the effectiveness of the proposed method through the ablation study and quantitative comparison on a range of textured meshes.

\begin{figure*}[htbp]
  \centering
   \includegraphics[width=1.0\linewidth]{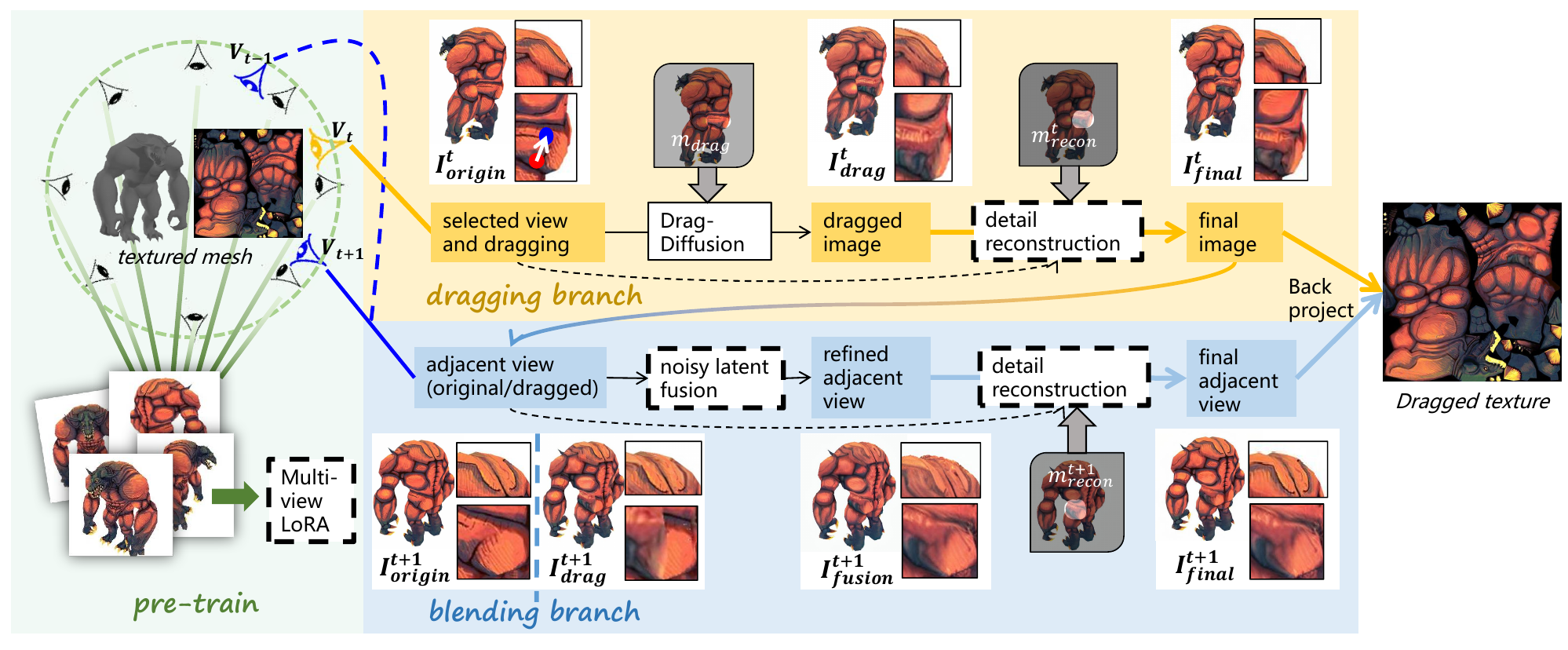}
   \caption{Overview of our method. Beginning with the multi-view LoRA pre-train, the dragged texture is generated via the dragging branch and blending branch. Our method involves optimizing the training strategy, fusion of noisy latent images, and reconstructing details outside the drag region (depicted by dashed boxes) to achieve the desired texture.}
   \label{fig:overview}
\end{figure*}

\section{Related Work}
\label{sec:formatting}

{\bf Text-to-Image generation.}\quad 2D AIGC has made significant strides prior to the advent of 3D content generation. Starting from GAN-based image generation, methods like StyleGAN \cite{karras2019style,karras2020analyzing} train a generator to produce an image that is indistinguishable from a real one. Recently, diffusion models have harnessed text embeddings to train a noise predictor. This predictor progressively denoises Gaussian noise, ultimately generating images that meet specified requirements \cite{ho2020denoising,song2020denoising,dhariwal2021diffusion}. Notable works in this domain include Imagen \cite{saharia2022photorealistic}, DALL-E \cite{ramesh2021zero,ramesh2022hierarchical}, GLIDE \cite{nichol2021glide}, and Stable Diffusion \cite{rombach2022high}. Among these, Stable Diffusion stands out by shifting the denoising process from the image space to the latent space, resulting in a significant reduction in computational overhead incurred during the diffusion process. Based on such models, conditional diffusion models have been proposed to support inputs like depth maps \cite{zhang2023adding}, coarse segmentation maps \cite{po2023compositional}, as well as other specific inputs. Besides, these models have been proven effective in addressing a wide range of image-to-image tasks like inpainting \cite{avrahami2022blended,avrahami2023blended}.

{\bf 2D image editing.}\quad Most of 2D image editing is based on text as input \cite{hertz2022prompt,brooks2023instructpix2pix,cao2023masactrl}, which is capable of modifying high-level semantic information rather than providing precise spatial control. The latter is typically realized by point-based editing accomplished through drag interactions. The method of DragGAN \cite{pan2023drag} performs motion supervision and point tracking in feature space, progressively moving the source point to the target point. FreeDrag \cite{ling2023freedrag} uses fuzzy localization and linear search to replace point tracking as well as updating the reference features of the source point during the dragging, aiming to avoid potential inaccuracies. DragDiffusion \cite{shi2023dragdiffusion} is similar to DragGAN, yet uses a diffusion model instead of GAN inversion and generation, aiming to improve the generalization of the model. DragonDiffusion \cite{mou2023dragondiffusion} also leverages the diffusion model to execute image dragging. It employs the original image generation process to guide image dragging through an attention mechanism and classifier guidance \cite{ho2022classifier}, ensuring the consistency of the appearance in the generated results.

{\bf Texture generation.}\quad The generation of textures on 3D models is also a prominent subject of AIGC. Previous methods like Text2Mesh \cite{michel2022text2mesh}, CLIP-Mesh \cite{mohammad2022clip}, GET3D \cite{gao2022get3d}, and DG3D \cite{zuo2023dg3d} are capable of generating both the 3D mesh and the associated texture. These processes are supervised by either the CLIP model \cite{radford2021learning} or GAN \cite{goodfellow2020generative}.
In the context of texture generation on a given mesh, Mesh2Tex \cite{bokhovkin2023mesh2tex} uses a hybrid neural field to bind the texture to the barycentric coordinate of the mesh surface. It then refines the resulting texture through optimization with the aid of an adversarial network. Meanwhile, Tango \cite{lei2022tango} generates the diffuse, specular, and roughness maps of the mesh using an SVBRDF network, all supervised by CLIP. Methods like Latent-Paint \cite{metzer2023latent} encode the texture image into a hidden space, generating a feature map from various viewpoints and refining the texture in the latent space through score distillation loss. Heuristic methods like TEXTure \cite{richardson2023texture} adapt existing text-to-image techniques directly to texture generation. It initiates the process from one view of the mesh and iteratively produces textures. Moreover, efforts have been made to improve global consistency in TexFusion \cite{cao2023texfusion} and \cite{tang2023text}. Text2Tex \cite{chen2023text2tex} further refines textures that may not have been generated optimally from certain angles.

{\bf 3D content editing.}\quad TextDeformer \cite{gao2023textdeformer} executes mesh deformation based on input text or sketch. Instruct-NeRF2NeRF \cite{haque2023instruct} and Instruct 3D-to-3D \cite{kamata2023instruct} operate with NeRF \cite{mildenhall2021nerf} and support user's text-based editing. Dragd3d \cite{xie2023dragd3d} extends drag-editing to 3D meshes, enabling the deformation of geometric components of 3D models through drag interactions.  
For texture editing on 3D models, as mentioned in TEXTure \cite{richardson2023texture}, the original texture is edited to a reasonable texture according to a text or scribble by adjusting the ``refine'' region in the model. ITEM3D \cite{liu2023item3d}, on the other hand, achieves text-based texture editing by optimizing the texture through text-based image editing on multi-view images. However, text-based texture editing may not provide precise spatial control over the texture. To address this limitation, we introduce DragTex, a generative model designed to facilitate interactive point-based editing of textures on 3D meshes.

\section{Method Overview}

We commence by presenting a framework overview of our method, as illustrated in Figure \ref{fig:overview}, to establish a context for the core algorithms detailed in the next section.
Starting with a textured mesh as input, our initial step involves the uniform sampling on the hemisphere encompassing the 3D mesh (black viewpoints in Figure \ref{fig:overview}) to gather a set of multi-view images. Prior to performing drag operations, we perform a fine-tuning process with a LoRA model applied to the UNet component of the diffusion model. Leveraging the collected multi-view images, we take this fine-tuning process as the ``pre-train'' part of our framework. 

Following the pre-train phase, our method affords the freedom to rotate the 3D mesh, allowing the user to select a specific viewpoint denoted as $V_t$ for texture dragging (\eg, the yellow viewpoint in Figure \ref{fig:overview}). Then, the rendered image $I_{origin}^t$ corresponding to the chosen viewpoint $V_t$, along with the user's specified source and target points for the drag operation, is fed into the fine-tuned 2D image-based dragging model \cite{shi2023dragdiffusion}. This step outputs the edited image $I_{drag}^t$, representing the effects of the drag operation. Subsequently, the original image $I_{origin}^t$, the edited image $I_{drag}^t$, and the reconstructed mask $m_{recon}^t$ drawn by users that aligns with the actual drag outcome, are collectively delivered into the ``detail reconstruction'' module. This module is responsible for producing the ultimate image $I_{final}^t$ of view $V_t$ and reinstating finer details outside of the drag-affected region. 

Furthermore, we select two neighboring views $V_{t-1}$ and $V_{t+1}$ (the blue viewpoints in Figure \ref{fig:overview}), positioned on both sides of view $V_t$. These views encompass the silhouettes of objects observed in viewpoint $V_t$ and are chosen for fusion. After inputting them into the ``noisy latent diffusion'' module, we get the neighboring viewpoint image \eg $I_{fusion}^{t+1}$ with the seam eliminated. Then, similar to the ``dragging branch'', we input the blended image $I_{fusion}^{t+1}$ into the ``detail reconstruction'' module to get the final result $I_{final}^{t+1}$. Finally, the edited images $I_{final}^{t-1}$, $I_{final}^{t}$, and $I_{final}^{t+1}$ are projected back onto the texture image, outputting the final texture following the drag operation.

\section{Generative Texture Editing}
\label{sec:method}

\subsection{Cross-view fusion refinement of drag regions}

The area near the silhouette of an object within a given view often undergoes obvious changes during the dragging process. If we directly project the dragged image back onto the atlas, any local content displacement in this region will lead to noticeable seams in neighboring views. To mitigate this issue, we address the need for fusion in the relevant regions of the affected neighboring views, aiming to eliminate these unwanted seams near the silhouette.
While traditional image blending methods, such as Poisson blending \cite{10.1145/1201775.882269} and latent optimization \cite{avrahami2023blended}, are effective for seamlessly integrating one image into another, they are typically geared towards scenarios where the goal is to maintain the content of two regions while ensuring smooth transitions. However, in our case, the contents of two regions become misaligned due to the dragging process's effect on one of them. Therefore, if we persist in maintaining the original content in both sides after blending, it will inevitably lead to a misalignment of the texture in the neighboring views.

Note that the discontinuity observed between both sides of the silhouette stems from the dragging operation in the preceding view, a process involving the optimization of the noisy latent image. It is a more rational strategy to rectify these discrepancies by merging the diffuse latent image and progressively refining it through the denoising process like \cite{richardson2023texture}. Thus, we will first optimize the texture $T$ using $I_{final}^{t}$ and render the dragged image $I_{drag}^{t'}$ in viewpoint $V_t'$ (equals to $V_{t+1}$ or $V_{t-1}$). Next, our strategy is to encode and noise $I_{drag}^{t'}$ and $I_{origin}^{t'}$ of adjacent view $V_t'$ to get the $Z_{origin}^{t'}(i)$ and $Z_{drag}^{t'}(i)$ at the $i$-th step of diffusion. Then we blend the latent image $Z_{drag}^{t'}(i)$ and $Z_{origin}^{t'}(i)$ to  $Z_{fusion}^{t'}(i)$, combining them by a mask $m_{fusion}$ that reflects the silhouette of the object of the view $V_t$ in the view $V_t'$. Concretely, we have
\begin{align}
    Z_{fusion}^{t'}(i) = \begin{cases}Z_{drag}^{t'}(i) \odot (1-m_{fusion})\\+Z_{origin}^{t'}(i) \odot m_{fusion}, & \ 0\leq i<20 \\\\
    Z_{fusion}^{t'}(i) \odot (1-m_{fusion})\\+Z_{origin}^{t'}(i) \odot m_{fusion}, & \ 20\leq i<40 
    \label{1}
\end{cases}
\end{align}
whereupon the fusion process is gradually completed during the denoising of $Z_{fusion}^{t'}(i)$.

Correspondingly, with $M_{LoRA}$ fine-tuned by multi-view images, we have
\begin{align}
    Z_{fusion}^{t'}(i-1)  = M_{LoRA} (Z_{fusion}^{t'}(i) )
\end{align}

In addition, given that our method revolves around progressive texture editing, it shares a limitation with texture generation through view iteration. Specifically, in cases where the surface normals deviate significantly from the viewing direction, the corresponding texture requires refinement. As a result, our $update\_mask$ and $checker\_mask$ are determined based on the normal maps as the method of TEXTure \cite{richardson2023texture}, to identify regions in the new view that require refinement and keep the consistency near the silhouette at the same time. In practice, we use
\begin{align}
    m_{fusion} = \begin{cases}checker\_mask, & \ 0 \leq i<20 \\
    update\_mask, & \ 20\leq i<40 
\end{cases}
\end{align}
where the $checker\_mask$ uses the checkerboard in the area around the silhouette while the $update\_mask$ divides the sides of the silhouette directly.

\subsection{Reconstruction of texture details in non-drag regions}

\begin{figure*}
  \centering
   \includegraphics[width=1.0\linewidth]{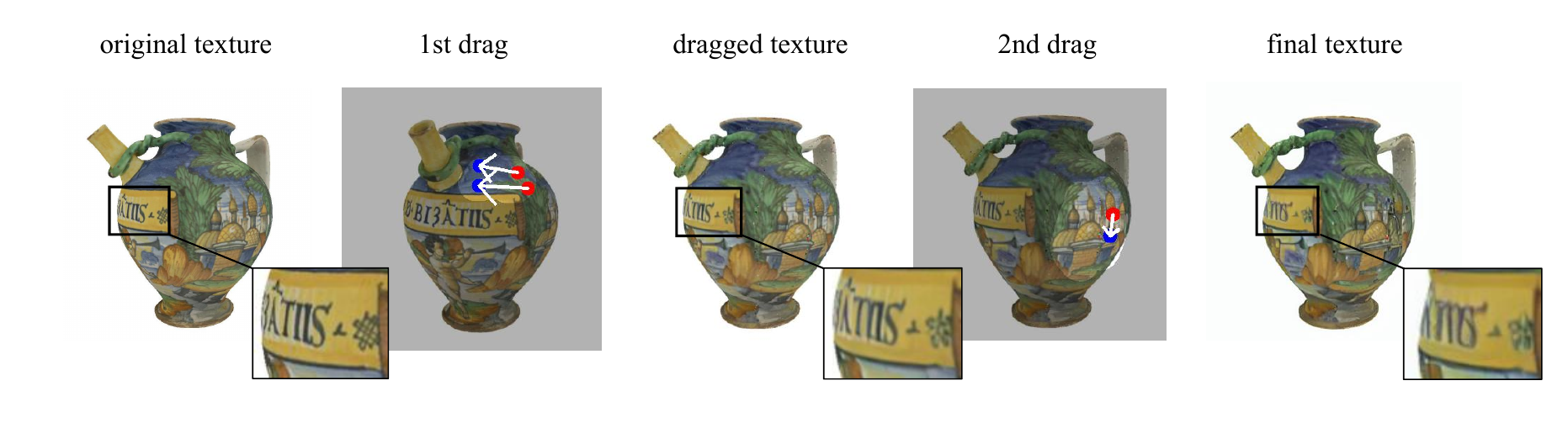}

   \caption{An example of reconstruction error accumulation. After the first drag in view-1, the region outside the drag region is not well reconstructed, as shown in view-2. Then, another drag in view-2 leads to more errors.}
   \label{fig:acc}
\end{figure*}

It is known that DragDiffusion builds upon the Stable Diffusion model \cite{rombach2022high}, employing a process that involves dragging features through the latent space, followed by decoding. However, the inherent limitation of the decoder in accurately representing high-frequency details results in a noticeable loss of fine image details. Even without any drag operation, such artifacts occur throughout the stages of encoding, diffusion, denoising, and decoding, as depicted in Figure \ref{fig:acc}. It becomes worse when generalizing the above 2D image generation process to textures on 3D meshes, as the details in the same place may be covered by multiple neighboring views participating in multiple drag interactions as non-drag regions and causing accumulation of error on the 3D mesh texture. As illustrated in Figure \ref{fig:acc}, the texture details degrade more substantially after dragging points from multiple related views.

In response to the above issue, we employ a fine-tuning process for the decoder, akin to the blended latent diffusion technique \cite{avrahami2023blended}. This fine-tuning process aims to restore details in regions outside the dragging and blending area, following the completion of the diffusion process. Concretely given the input images $I_{origin}$, $I'$ and mask $m_{recon}$, initializing the latent code ${z_0}$, we fine-tune the decoder $D_{\theta}$ with the following formula \cite{avrahami2023blended}: 

\begin{footnotesize}
\begin{flalign}
&\theta^*=\underset{\theta}{\operatorname{argmin}}\left\|D_\theta\left(z_0\right) \odot m_{recon}-I' \odot m_{recon}\right\|+\nonumber\\
&\lambda_{recon}\left\|D_\theta\left(z_0\right) \odot(1-m_{recon})-I_{origin} \odot(1-m_{recon})\right\|
\end{flalign}
\end{footnotesize}
Then, we use the above weights to achieve detailed reconstruction by computing
\begin{align}
I_{final} = D_\theta^*\left(z_0\right)
\end{align}

It is worth noting that DragDiffusion does not strictly preserve the current view outside the ``drag mask'' $m_{drag}$. Meanwhile, the affected regions of neighboring views after blending are uncertain. Acknowledging such two uncertainties, we allow users to create a ``reconstruction mask'' $m_{recon}$ which is different from $m_{drag}$. This mask should encompass the regions of the texture affected by the dragging or blending, ensuring the preservation of details in both the dragged or fused region and remaining original texture. Specifically, for the ``dragging branch'' as shown in Figure \ref{fig:overview}, we take $I_{drag}^{t}$ as $I'$, and draw $m_{recon}^{t}$ based on the actual drag outcome as inputs, intending to get detailed $I_{final}^{t}$. For the ``blending branch'', we utilize $I_{fusion}^{t'}$ as $I'$ and draw $m_{recon}^{t'}$ based on the region affected by blending, enabling the reconstruction of $I_{final}^{t'}$.

\subsection{Training with multi-view images}

In order to meet the need for freely editing textures, it is clear that the ability of free-view drag is crucial. In the framework of our previously outlined approach, LoRA training is required each time a texture is dragged in the selected view using DragDiffusion. Moreover, when combining neighboring views, additional training for LoRA is required. With two neighboring views involved in this process, the time complexity of training LoRA for each dragging interaction scales to an order of $3n$ for $n$ drag interactions. This level of computational overhead is unsatisfying.

To expedite the training process, we choose an alternative scheme that involves pre-training a multi-view LoRA before engaging in the dragging and blending. We initially render images from the surrounding views evenly distributed around the 3D mesh. We then leverage LoRA, a fine-tuning model capable of training on multiple images with similar styles and subsequently reconstructing them. By following this scheme, we can train the model in advance, obviating the need to retrain it for each drag interaction. As a result, it can also significantly enhance interaction efficiency.

\begin{figure}[htbp]
  \centering
   \includegraphics[width=1.0\linewidth]{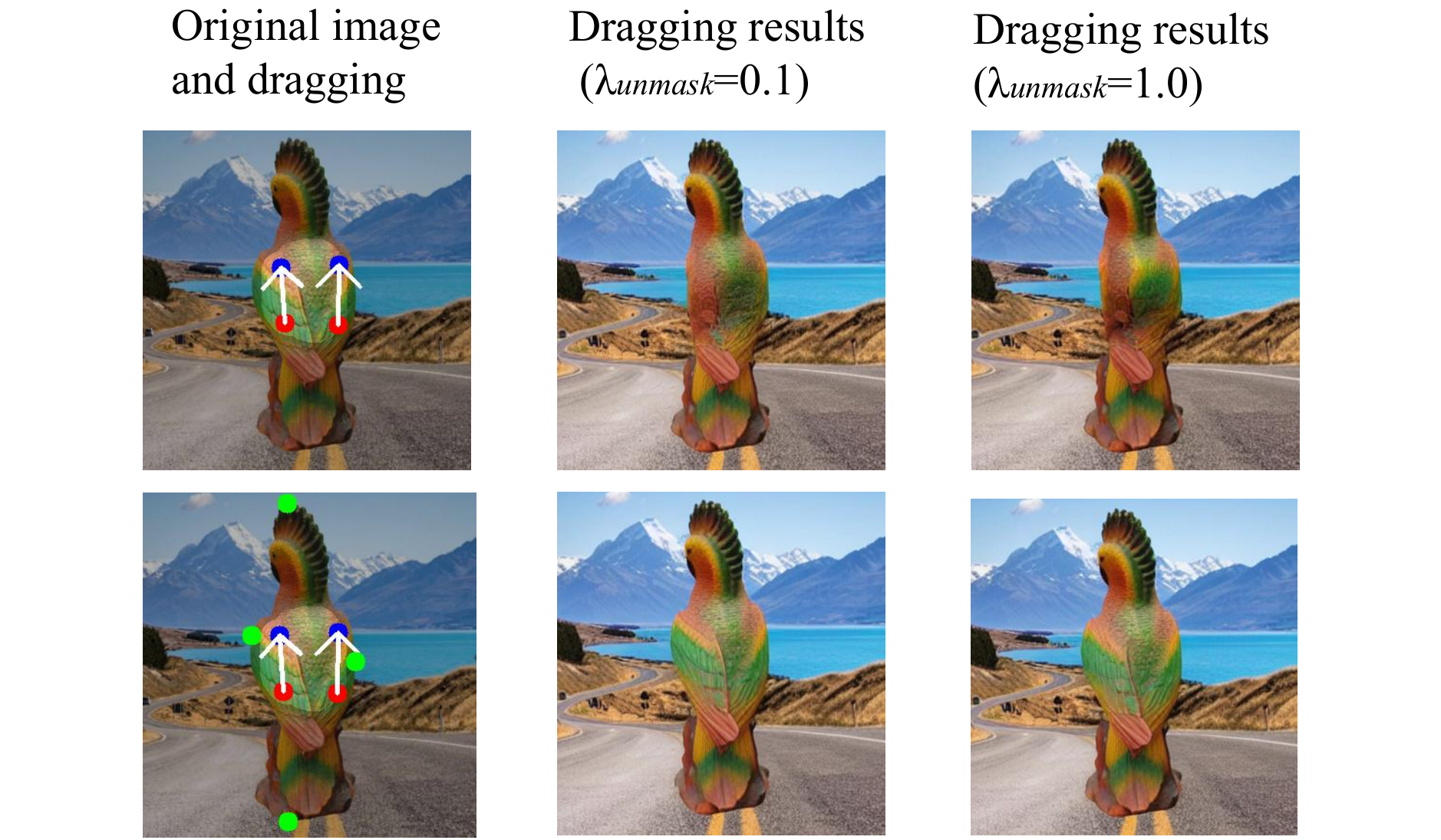}

   \caption{The results of adding static control points (green points in the second row). }
   \label{fig:static}
\end{figure}
\begin{figure*}
  \centering
   \includegraphics[width=1.0\linewidth]{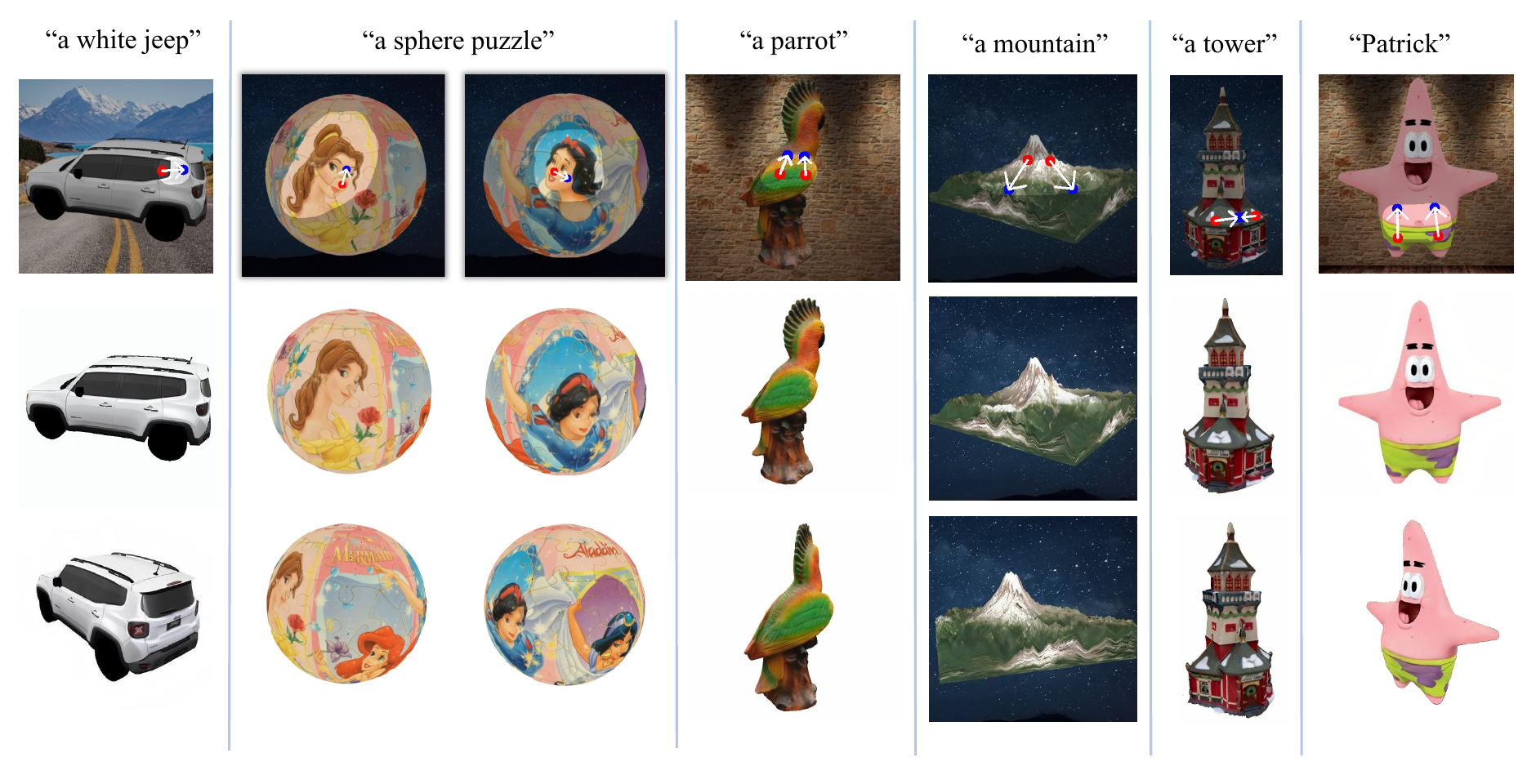}

   \caption{Our results on different kinds of textured meshes. The first row shows the original images of selected views and users' drag, and the second and third row shows the rendered images of the dragged texture in the selected view and another related view.}
   \label{fig:finalres}
\end{figure*}

\subsection{Adding static control points}
Even though existing methods \cite{pan2023drag,shi2023dragdiffusion} add relevant constraints to guarantee that the region outside the mask does not change, dragging based on motion supervision is still a lossy optimization problem. For example, in practice, we have found that dragging inside the object can individually lead to changes in its geometry. This would result in part of the image's background being incorrectly back-projected onto the object's texture atlas, which is unacceptable. To handle this problem, we propose a simple yet effective trick that involves calculating some points on the silhouette of the object as static control points. Using the farthest point sampling method which iteratively selects the point with the largest sum of distances from existing sampling points, we adjust the loss fusion in motion supervision \cite{shi2023dragdiffusion}. This adjustment not only constrains the features at the dragged point to move toward the target point and keeps the region outside the dragging mask unchanged (controlled by weight $\lambda_{unmask}$), but also emphatically constrains the features at the static point to remain at the origin, thus effectively maintains the geometry (see Figure \ref{fig:static}).

\begin{figure*}
  \centering
   \includegraphics[width=0.95\linewidth]{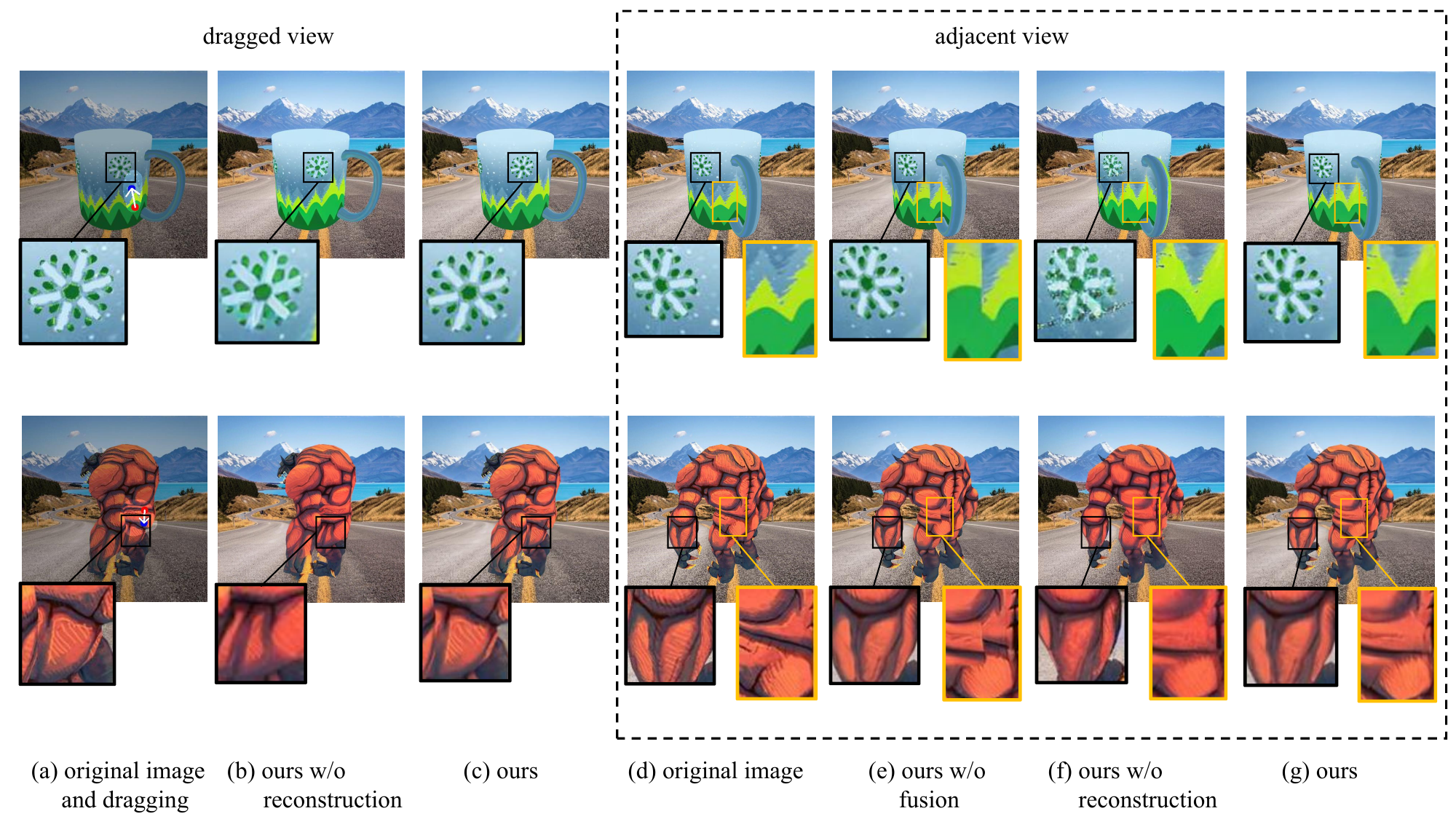}

   \caption{The results of ablation study. The dragged region is indicated with a yellow border and the non-drag region is with a black border.}
   \label{fig:ablation}
\end{figure*}

\section{Experiments}

We have implemented our method and conducted extensive testing with various drag interactions.
The textured meshes used for testing are obtained from a range of sources, like TurboSquid~\footnote{https://www.turbosquid.com}, Sketchfab~\footnote{https://sketchfab.com}, Objaverse \cite{deitke2023objaverse}, and others, including eight different kinds of objects.
We employed Stable Diffusion v1-5 from the DragDiffusion pipeline. When utilizing drag in the latent space and fusion, we configured 50 steps for DDIM and fusion, and 35 steps for dragging. Similarly, we maintained a rank of 16 and a learning rate of $2\times10^{-4}$ in LoRA training. For single-view training, the number of training steps was set to 200. In multi-view training, we rendered 10 images uniformly sampled on a hemisphere with a fixed radius as a training set and set 1000 steps as the default, based on the prior training of LoRA. Next, we elaborate on the details of our experiments.

\subsection{Results}

We employ the pre-trained model on multi-view images to execute the dragging, cross-view fusion, and details reconstruction, resulting in the final editing outcomes. It is evident that our fusion and reconstruction perform admirably when utilizing the multi-view image pre-trained model. As shown in Figure \ref{fig:finalres}, the textures after dragging exhibit continuity across cross-view regions while effectively preserving fine details. For visual demonstrations of these features, we refer the reader to the companion video.

A set of textured meshes including cars, spherical puzzles, parrots, mountains, furniture, buildings, and cartoons all generate drag-compliant, high-quality textures after dragging with our pipeline. For example, the details of flowers and letters in the texture of ``a sphere puzzle'' are well preserved, as are the details in the other textures. When dragging across perspectives involving \eg Patrick's stretched pants, our model eliminates the appearance of unnatural seams in the adjacent view. It takes about 20 seconds to drag or blend with an A100 GPU. We then got ``a mountain'' after a snowy day, ``a jeep'' with longer windows, ``a parrot'' with more green feathers, \etc.
\subsection{Ablation study}

To examine the individual impact of cross-view fusion and detailed reconstruction on the final edited textures, we conducted an ablation study by disabling one of them at a time. Figure \ref{fig:ablation} illustrates some of the results from this ablation study. It can be seen when there is no cross-view fusion, noticeable seams become apparent in the adjacent views of the drag operation, leading to inconsistent textures. Conversely, when detailed reconstruction is omitted, the texture exhibits significant distortion in areas without dragging, resulting in a substantial degradation in quality. However, when both cross-view fusion and detailed reconstruction are employed, our method successfully maintains the dragging effect while ensuring that the remaining details have consistency with the original texture. This highlights the significance of both components in achieving high-quality edited textures.

\subsection{Comparison of different training strategies}

In order to avoid the time overhead associated with training LoRA for each
view in DragDiffusion, we train LoRA with multi-view images in advance. Considering that training images from other views may affect the reconstruction of the current view image, we also need to include another scheme in the comparison, \ie, the DragDiffusion with Masactrl \cite{cao2023masactrl} guiding yet without LoRA fine-tuning. In order to find out the acceptable and superior alternative scheme, we evaluate the effectiveness of the above two strategies: the Drag with Masactrl and the Drag with Multi-view LoRA with Masactrl, based on both qualitative and quantitative experiments. 

In previous work, the necessity of using single-view LoRA has already been demonstrated by quantitative comparisons. Therefore, in order to measure the relative generative quality of the multi-view image training strategy, we do the following two quantitative experiments with the baseline of the single-view image training strategy. Firstly, we calculate and compare the PSNR, SSIM, and LPIPS \cite{zhang2018unreasonable} between the two alternative strategies and the baseline respectively to find out the strategy that can show comparable results. Moreover, we also counted the percentage of ``Failure cases'' that failed to maintain the original image content in all the tests as a reference. Secondly, we measure the mean distance (MD) like \cite{pan2023drag,shi2023dragdiffusion} to evaluate the effectiveness of dragging among these strategies.

\begin{figure}
  \centering
   \includegraphics[width=1.0\linewidth]{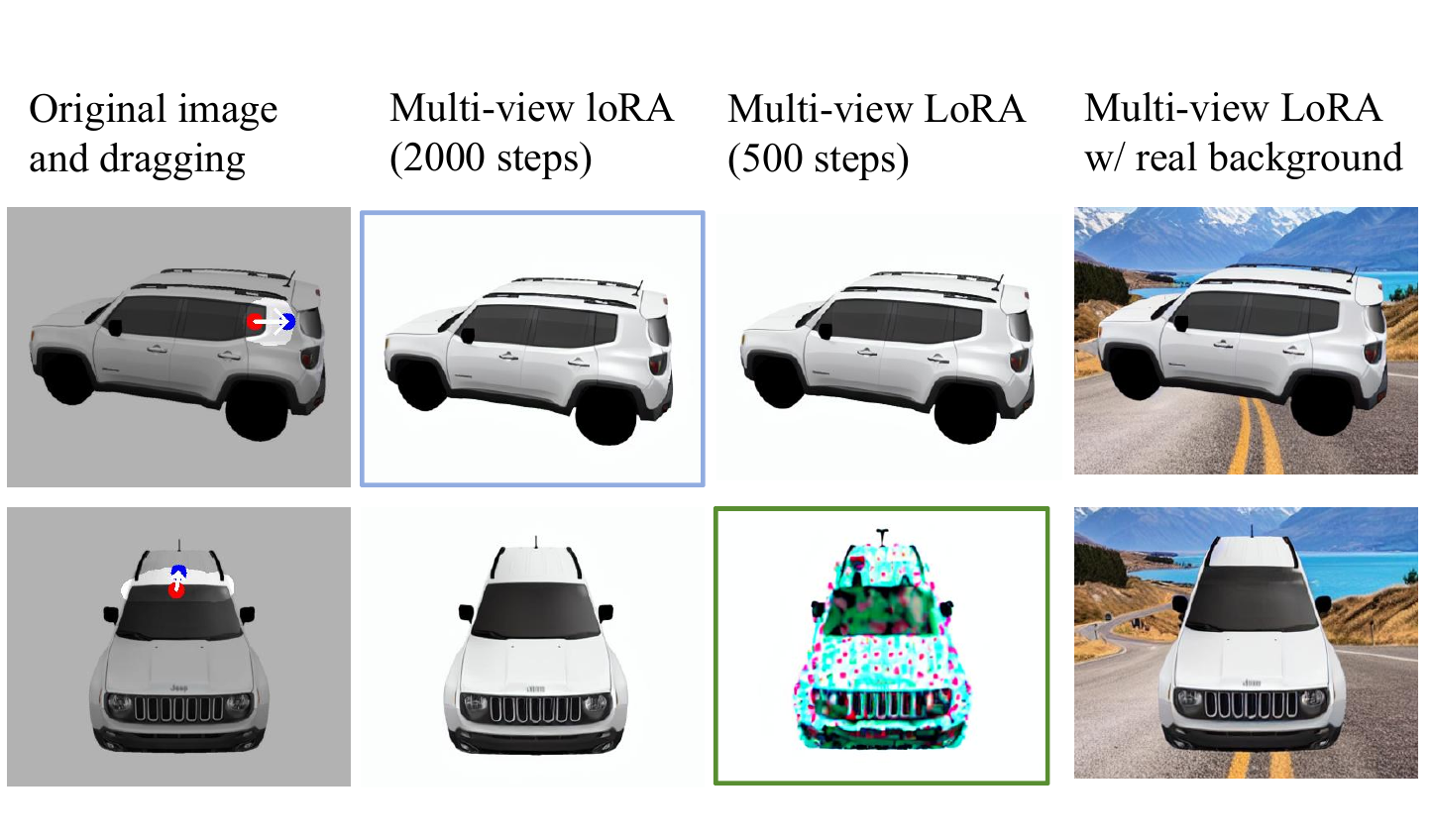}

   \caption{Two middle columns show the trade-off of training with all-white background images, where the results of the over-fitting failure and reconstruction failure are shown with blue border and green border. The last column shows the results of training LoRA with real background images.}
   \label{fig:white_bg}
\end{figure}

It is observed that simply using all-white backgrounds to render images can result in reconstruction failure, as shown in Figure \ref{fig:white_bg}. Images of some view have the over-fitting problem with more training steps, while some images are reconstructed incorrectly with fewer training steps, leading to an unsatisfying trade-off. This is because synthetic images with pure white backgrounds are beyond the scope of the diffusion model's large training set of real images, which causes a degradation of the generative ability. Therefore, we use real images as the background for mesh rendering in an attempt to eliminate the above trade-off, and the results are expected as shown in Figure \ref{fig:white_bg}.

\begin{figure}
  \centering
   \includegraphics[width=1.0\linewidth]{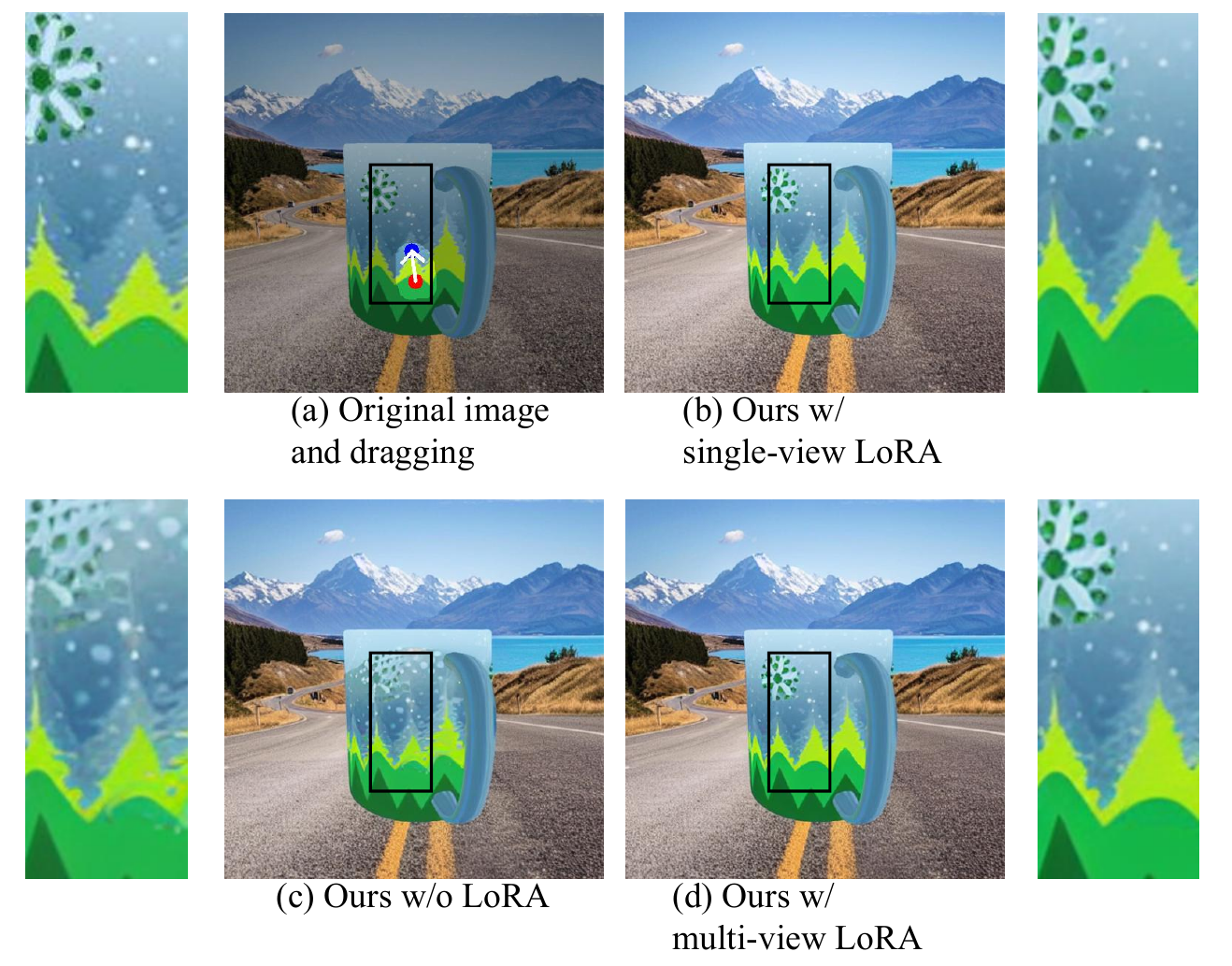}

   \caption{Qualitative comparison of different training strategies.}
   \label{fig:comparison}
\end{figure}

With the above adjustment, we organize our dataset with 4 different backgrounds to make sure users can find a suitable one for their model. For every object, we randomly pick 2-5 views in or out of the training set as a testing set. Meanwhile, we use all the background to test all the models to ensure that our method has convincing robustness. The statistics in Table \ref{tab:example}, Table \ref{tab:MD}, and the qualitative results in Figure \ref{fig:comparison} show that training LoRA with multi-view images has a significant effect on the editing quality. Additionally, the results of multi-view LoRA training are close to those of single-view LoRA training and therefore meet our needs. Consequently, we train multi-view images as a pre-training part before all dragging operations.

For texture dragging with free viewpoints, directly applying 2D image dragging requires training the LoRA model for each dragging viewpoint, which means that the time complexity of training for $n$ dragging viewpoints plus their two neighboring viewpoints is $3n$. However, our approach uses five times the training time of a single image (at a constant level) as pre-training, which eliminates the need to train before each dragging and provides the user with convenience. 

\begin{table}
  \centering
  \small
  \begin{tabular}{l|c|c|c|c}
    \toprule
    Method & PSNR$\uparrow$ & SSIM$\uparrow$ & LPIPS$\downarrow$ & Failure Cases\\
    \midrule
    Ours w/o LoRA & 26.84 & 0.87 & 0.13 & 28.2 \% \\
    \makecell{Ours w/ multi-\\view LoRA} & {\bf 31.30} & {\bf 0.94} & {\bf 0.03} & {\bf 4.6 \%}\\
    \bottomrule
  \end{tabular}
  \caption{Quantitative comparison between two efficient alternative strategies.}
  \label{tab:example}
\end{table}

\begin{table}
  \centering
  \small
  \begin{tabular}{l|c|c|c}
    \toprule
    Method & \makecell{Ours w/ multi-\\view LoRA} & Ours w/o LoRA & \makecell{Ours w/ single-\\view LoRA}\\
    \midrule
    MD$\downarrow$ & 33.76 & 41.53 & 32.36 \\
    \bottomrule
  \end{tabular}
  \caption{The mean distance (MD) of different strategies.}
  \label{tab:MD}
\end{table}

\section{Conclusion}

We have proposed a new task of texture dragging on 3D mesh and presented a generative solution, namely \emph{DragTex}, which excels in facilitating point-based editing while preserving the intricate details of the original texture. Our method is rooted in the principles of the diffusion model and trains LoRA with multi-view images before engaging in any drag interactions, thus optimizing and expediting the entire process. After freely selecting a view and dragging, we blend the noisy latent image on the overlapping region between two views and meanwhile fine-tune a decoder to make the edited texture consistent and elaborate. Extensive qualitative and quantitative experiments have shown the effectiveness of our method. 

As the future work, we intend to optimize the inference process to facilitate real-time feedback during drag interactions. Besides, it is promising to extend point-based texture dragging to a broader spectrum of interaction types, \eg, strokes or free-hand drawing on 3D meshes.

{
    \small
    \bibliographystyle{ieeenat_fullname}
    \bibliography{main}
}


\end{document}